# Phase Transition Frequency as a Training-Time Predictor of Test Accuracy in ResNets

Arunan Jithendrakumar[1*]

[1*]Independent Researcher, Chennai, Tamil Nadu, 600080, India.

Corresponding author(s). E-mail(s): arunanj2005@gmail.com;

**Abstract**

The number of discrete class-separability jumps observed during ResNet fine-tuning is examined empirically as a predictor of final test accuracy. Across 75 experiments spanning four benchmarks (CIFAR-10, CIFAR-100, TinyImageNet, and CIFAR-10-C) and three architectures (ResNet-18, ResNet-50, and ResNet-101) with five to ten seeds per configuration, a strong within-dataset negative correlation is obtained on standard i.i.d. classification benchmarks: $\boldsymbol{r = -0.84}$ on CIFAR-10 ($\boldsymbol{p < 10^{-8}}$, $\boldsymbol{n = 30}$) and $\boldsymbol{r = -0.87}$ on CIFAR-100 ($\boldsymbol{p < 10^{-5}}$, $\boldsymbol{n = 15}$). Under distributional stress the relationship attenuates: TinyImageNet yields $\boldsymbol{r = -0.45}$ and the CIFAR-10-C corruption benchmark yields $\boldsymbol{r = -0.19}$. Two additional analyses discipline the empirical claim. A partial correlation controlling for architecture depth (treated as a linear covariate) shows that on CIFAR-100 the transition count retains statistically significant predictive power ($\boldsymbol{r_{\text{partial}} = -0.69}$, $\boldsymbol{p = 0.007}$); the corresponding result under the stricter categorical conditioning is not established at $\boldsymbol{n = 15}$. A comparison against six alternative training-curve signals shows that transition count achieved the strongest correlation among the evaluated signals on CIFAR-100 and one of the strongest on CIFAR-10, but is dominated by other signals on the two stressed benchmarks. The comparison is restricted to training-curve-level signals; comparisons against effective rank, Hessian sharpness, Fisher information, margin, and neural-collapse measures, which are the strongest competitors in the current literature, are not part of the present study and remain open. The observation is presented as an in-distribution training-quality probe among a family of candidate probes, and an inexpensive detection procedure suitable for logging alongside a standard training loop is provided.

# 1 Introduction

The gap between the training loss a neural network is able to fit and the test loss it eventually achieves is one of the central puzzles of supervised deep learning. Beyond raw capacity and data volume, the dynamics of training itself, meaning the pattern by which internal representations reorganise across epochs, plausibly carries information about how well the eventual solution will generalise. This paper measures one particular aspect of those dynamics: the number of discrete moments during training at which the class geometry of internal representations reorganises. These moments are termed *phase transitions* throughout the paper and are counted via an epoch-to-epoch threshold on class separability. The metric is inexpensive to compute alongside a standard training loop.

The empirical finding is summarised in Figure 1 and consists of a two-regime pattern. On CIFAR-10 and CIFAR-100, the number of transitions is a strong negative predictor of final test accuracy within each dataset ($r \leq -0.83$, $p < 10^{-5}$). On TinyImageNet (a harder task with 200 classes) and on CIFAR-10-C (a corruption robustness benchmark), the within-dataset correlation attenuates substantially, and on CIFAR-10-C the architectural ordering between ResNet-18 and ResNet-50 is even mildly inverted despite the large difference in transition count. Two follow-up analyses (Sections 4.3 and 4.4) show that on CIFAR-100 the transition count retains predictive power after controlling for architecture depth, but that on CIFAR-10 depth alone is a somewhat stronger single-variable predictor, and that on the two stressed benchmarks neither transition count nor any of six alternative training-curve signals achieves a strong association with test accuracy. Together, these results support a modest but clear reading: the metric is a useful in-distribution training-quality probe rather than a universal predictor of accuracy under distribution shift.

The scope of this study is limited to three ResNet variants on four image-classification benchmarks. Vision Transformers, MLP-Mixers, and non-vision modalities were not evaluated (Section 6).

# 2 Related Work

## 2.1 Information Bottleneck

The Information Bottleneck principle Tishby and Zaslavsky (2015) posits that a neural network compresses its input while preserving task-relevant information. Its interpretation in the context of deep networks has been contested Shwartz-Ziv and Tishby (2017); Saxe et al. (2018). The measurement made here does not attempt to estimate input–representation mutual information; the focus is on how the internal class geometry of representations reorganises during training.

## 2.2 Phase Transitions and Sudden Learning

Phase transitions in learning have been reported at various scales. Fort and Ganguli (2019) identified sudden discrete jumps in loss during SGD, Spigler et al. (2019) described jamming behaviour in loss landscapes, and Achille et al. (2019) studied critical periods early in training. Power et al. (2022) demonstrated delayed sudden generalisation on algorithmic tasks. The transitions counted here are distinct from grokking: rather than a delayed generalisation jump on a fixed dataset, they are within-training reorganisations detected by an epoch-to-epoch change in class separability, and it is their aggregate count that is examined against final accuracy.

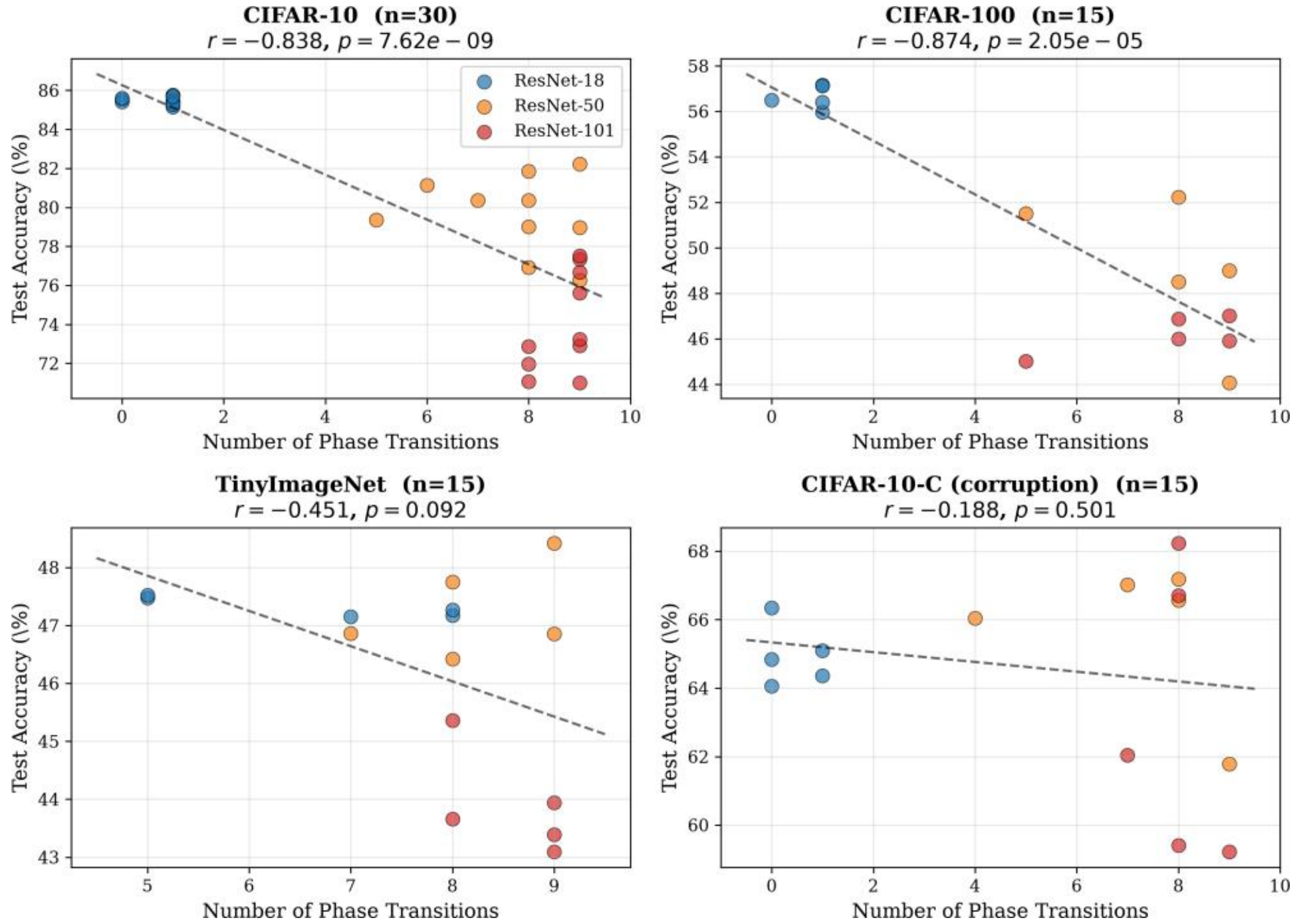

**Fig. 1** Test accuracy against number of phase transitions, for the three ResNet architectures on each of the four benchmarks. Within-dataset regression fits appear as dashed lines. The negative correlation is strong on CIFAR-10 and CIFAR-100 ($r \leq -0.84$, $p < 10^{-5}$) and attenuates as task difficulty and distribution shift increase.

## 2.3 Training Dynamics, Neural Collapse, and Generalisation

Several strands of prior work analyse how the trajectory taken by SGD relates to final performance. Lewkowycz et al. (2020) characterised feature learning under varying learning-rate regimes; Jastrzebski et al. (2017) analysed factors influencing the minima reached by SGD; and neural collapse describes the convergence of feature geometry at the end of training Papyan et al. (2020). The metric proposed here is complementary in that it counts discrete reorganisation events during training rather than characterising the endpoint geometry. Section 4.4 compares its predictive power against several training-curve-derived alternatives on the same experimental grid.

## 2.4 Robustness and Distribution Shift

Robustness under distribution shift has been examined extensively through benchmarks such as CIFAR-10-C Hendrycks and Dietterich (2019), ImageNet-C, ImageNet-R, and ImageNet-A Hendrycks et al. (2021); Recht et al. (2019); Taori et al. (2020). Prior work has generally characterised robustness gaps and their determinants without directly connecting them to training-time reorganisation signals. The CIFAR-10-C result reported here, that the transition–accuracy correlation attenuates under corruption, is consistent with the view that robustness under distribution shift is governed by mechanisms that in-distribution training dynamics do not capture.

# 3 Methodology

The experimental grid uses class separability, computed by the Fisher criterion, as the underlying signal from which phase transitions are detected. A frequency-domain framework that motivates the choice of separability as a proxy is deferred to Appendix A; none of the empirical results below depends on it.

## 3.1 Detection Signal

For each monitored layer $\ell$ and evaluation epoch $t$, three inexpensive proxies are computed from $\boldsymbol{A}_\ell$ over a random subset of $N = 1000$ training samples: the representation magnitude, the effective rank, and the class separability. All three were tracked during preliminary work; class separability was found to be the most sensitive to reorganisation events and is used as the detection signal throughout. The other two are retained as auxiliary diagnostics.

The class separability is defined via the Fisher criterion,

$$\mathrm{Sep}_\ell = \frac{\mathrm{Tr}(\boldsymbol{S}_B)}{\mathrm{Tr}(\boldsymbol{S}_W) + \mathrm{Tr}(\boldsymbol{S}_B)}, \tag{1}$$

where $\boldsymbol{S}_W$ and $\boldsymbol{S}_B$ are the within-class and between-class scatter matrices respectively. A naive evaluation of $\mathrm{Tr}(\boldsymbol{S}_B)$ and $\mathrm{Tr}(\boldsymbol{S}_W)$ materialises a $[d_\ell, d_\ell]$ scatter matrix per class; for ResNet-50 early layers with $d_\ell \approx 16{,}384$ and benchmarks with 100 or more classes, the naive computation exceeds 100 GB of host memory and made several cells of the experimental grid infeasible on the cloud instances used. Since only the traces are needed, the identities

$$\mathrm{Tr}(\boldsymbol{S}_B) = \sum_c n_c \|\boldsymbol{\mu}_c - \boldsymbol{\mu}\|^2, \tag{2}$$

$$\mathrm{Tr}(\boldsymbol{S}_W) = \sum_c \frac{1}{n_c - 1} \sum_{\boldsymbol{x} \in c} \|\boldsymbol{x} - \boldsymbol{\mu}_c\|^2 \tag{3}$$

are used instead. The resulting computation is $O(N \cdot d_\ell)$ in both time and memory rather than $O(K \cdot d_\ell^2)$ and produces numerically identical trace values (verified against the naive computation). This optimisation is what made the TinyImageNet cells (200 classes) feasible in practice.

### 3.2 Phase Transition Detection

Transitions are detected on the class-separability trace by Algorithm 1.

**Algorithm 1** Phase transition detection.

1: **Input:** separability $\{S_\ell^{(t)}\}$ for epochs $t = 1, \ldots, T$ and layers $\ell = 1, \ldots, L$; threshold $\tau$.
2: **Output:** set of transition epochs $\mathcal{T}$.
3: $\mathcal{T} \leftarrow \emptyset$
4: **for** $t = 2$ **to** $T$ **do**
5:     **for** $\ell = 1$ **to** $L$ **do**
6:         **if** $|S_\ell^{(t)} - S_\ell^{(t-1)}| / S_\ell^{(0)} > \tau$ **then**
7:             $\mathcal{T} \leftarrow \mathcal{T} \cup \{t\}$; **break**
8:         **end if**
9:     **end for**
10: **end for**

The threshold $\tau = 0.05$ (a 5% relative change from the epoch-zero baseline) is used throughout. A full threshold sweep was not conducted in this study; a rank-based robustness proxy is reported in Section 4.4 and the limitation is discussed further in Section 6.

### 3.3 Experimental Setup

Three architectures are used: ResNet-18, ResNet-50, and ResNet-101 He et al. (2016). All are initialised with ImageNet-pretrained weights and fine-tuned on the target dataset, with the final classifier head replaced to match the target class count. Four datasets are used. CIFAR-10 Krizhevsky et al. (2009) and CIFAR-100 Krizhevsky et al. (2009) test in-distribution behaviour at differing task difficulty; TinyImageNet (200 classes at $64 \times 64$) provides a coarser ImageNet-style benchmark that remains tractable at grid scale; and CIFAR-10-C Hendrycks and Dietterich (2019) tests corruption robustness, with models trained on standard CIFAR-10 and evaluated on a corrupted test set at severity 5 averaged across four corruption types (Gaussian noise, shot noise, defocus blur, and frost).

Optimisation uses SGD with momentum 0.9, initial learning rate 0.01 with cosine annealing over 50 epochs, weight decay $5 \times 10^{-4}$, and batch size 64. Ten seeds are used per configuration on CIFAR-10 (a legacy of the earlier exploratory phase of this work); five seeds are used on each of CIFAR-100, TinyImageNet, and CIFAR-10-C. The total experimental grid comprises 75 cells. All four ResNet blocks are monitored, and separability is computed every 5 epochs. Experiments were run as isolated cloud training jobs, one per (dataset, architecture) pair, on instances providing a single NVIDIA T4 GPU with 128 GB host RAM. Within each job, seeds were run sequentially in fresh Python processes to prevent memory accumulation across runs. Total wall-clock time across the grid was approximately 60 hours.

# 4 Results

## 4.1 Aggregate

Table 1 summarises the full experimental grid.

**Table 1** Full experimental grid ($n$ = 75). Each cell reports the mean ± standard deviation of the number of phase transitions (#$T$) and the test accuracy in percent, over the number of seeds indicated.

| Dataset | Metric | ResNet-18 | ResNet-50 | ResNet-101 | Seeds/cell |
|---|---|---|---|---|---|
| CIFAR-10 | #$T$ | 0.70 ± 0.48 | 7.70 ± 1.34 | 8.70 ± 0.48 | 10 |
| | Acc | 85.50 ± 0.21 | 79.64 ± 1.96 | 74.02 ± 2.54 | |
| CIFAR-100 | #$T$ | 0.80 ± 0.45 | 7.80 ± 1.64 | 7.80 ± 1.64 | 5 |
| | Acc | 56.63 ± 0.51 | 49.06 ± 3.21 | 46.16 ± 0.81 | |
| TinyImageNet | #$T$ | 6.60 ± 1.52 | 8.20 ± 0.84 | 8.60 ± 0.55 | 5 |
| | Acc | 47.32 ± 0.17 | 47.26 ± 0.81 | 43.89 ± 0.88 | |
| CIFAR-10-C | #$T$ | 0.40 ± 0.55 | 7.20 ± 1.92 | 8.00 ± 0.71 | 5 |
| | Acc | 64.94 ± 0.88 | 65.72 ± 2.24 | 63.12 ± 4.16 | |

Pooled across all 75 experiments the raw Pearson correlation between transition count and test accuracy is $r = -0.33$ ($p = 3.6 \times 10^{-3}$). This pooled statistic underestimates the within-dataset relationship, since the four benchmarks have very different accuracy baselines and the between-dataset variance dilutes the within-dataset signal. The dataset-conditional analysis reported next is the appropriate estimand.

## 4.2 Within-Dataset Correlation

Within-dataset Pearson correlations, together with Spearman rank correlations as a monotone-transformation-invariant sanity check, are reported in Table 2.

**Table 2** Within-dataset correlation between phase transition count and test accuracy. The Spearman rank correlation is invariant under monotone transformations of the transition-count signal and serves as a partial robustness check against the specific choice of detection threshold.

| Dataset | $n$ | Pearson $r$ | $p$-value | Spearman $\rho$ | Regime |
|---|---|---|---|---|---|
| CIFAR-10 | 30 | −0.838 | $< 10^{-8}$ | −0.782 | In-distribution |
| CIFAR-100 | 15 | −0.874 | $< 10^{-5}$ | −0.748 | In-distribution |
| TinyImageNet | 15 | −0.451 | 0.092 | −0.447 | Harder task |
| CIFAR-10-C | 15 | −0.188 | 0.501 | −0.071 | Distribution shift |

The Pearson and Spearman estimates agree on the qualitative pattern in every case: a strong negative association on the two in-distribution benchmarks, an attenuated one on TinyImageNet, and no significant association on CIFAR-10-C. Since Spearman is invariant to any strictly monotone transformation of the transition-count metric, its close agreement with Pearson indicates that the within-dataset ranking of experiments by transition count is not an artefact of the specific detection threshold $\tau = 0.05$: any threshold choice that preserves the ranking would yield the same conclusion.

The bootstrap 95% confidence intervals on within-dataset Pearson $r$ are $[-0.91, -0.72]$ on CIFAR-10, $[-0.97, -0.68]$ on CIFAR-100, $[-0.81, +0.06]$ on TinyImageNet, and $[-0.62, +0.32]$ on CIFAR-10-C. The two in-distribution intervals exclude zero clearly; the two stressed intervals do not.

## 4.3 Partial Correlation Controlling for Architecture

Because architecture is itself strongly correlated with test accuracy in this experimental design (see Section 4.4), a natural objection is that the transition–accuracy correlation may simply reflect an architecture-size effect. Two versions of this control are reported in Table 3. The first treats architecture as a linear variable in depth (values 18, 50, 101), which is a strong parametric assumption but statistically efficient. The second treats architecture as an unordered categorical variable (two dummy indicators, with ResNet-18 as reference), which is more conservative in that it makes no ordering assumption. Both are reported because they answer subtly different questions.

**Table 3** Partial Pearson correlation between transition count and test accuracy, controlling for architecture. Column "linear depth" encodes architecture as its layer count (18/50/101); column "categorical" uses two dummy variables for ResNet-50 and ResNet-101 with ResNet-18 as reference. Because the three architectures produce sharply different mean transition counts, the categorical variant is very demanding: only the within-cell variation is available to correlate with within-cell accuracy variation.

| | | | Linear depth | | Categorical |
|---|---|---|---|---|---|
| Dataset | $n$ | $r$ (raw) | $r_{\text{partial}}$ | $p$ | $r_{\text{partial}}$ ($p$) |
| CIFAR-10 | 30 | −0.838 | −0.308 | 0.10 | +0.042 (0.83) |
| CIFAR-100 | 15 | −0.874 | −0.687 | 0.007 | −0.259 (0.35) |
| TinyImageNet | 15 | −0.451 | +0.186 | 0.52 | −0.114 (0.69) |
| CIFAR-10-C | 15 | −0.188 | +0.110 | 0.71 | −0.239 (0.39) |

Under linear-depth conditioning, transition count retains statistically significant predictive power on CIFAR-100 ($r_{\text{partial}} = -0.687$, $p = 0.007$) after depth has been accounted for. The corresponding multiple regression on CIFAR-100 gives $\beta_T = -0.70$ ($p = 0.007$) and $\beta_D = -0.066$ ($p = 0.011$), with a combined $R^2 = 0.87$. Bootstrap 95% confidence intervals over 10,000 resamples give $\beta_T \in [-1.34, -0.03]$; the interval excludes zero but does so narrowly. This fragility is not surprising at $n = 15$, and the caveat should not be glossed: the CIFAR-100 result is the sharpest in the paper but rests on a modest sample.

Under categorical conditioning the result on CIFAR-100 weakens to $r_{\text{partial}} = -0.259$ ($p = 0.35$). The mechanical reason is that transitions vary sharply between architectures on CIFAR-100 (0.8 for ResNet-18 against 7.8 for both ResNet-50 and ResNet-101), so once the between-architecture component is removed the residual within-cell variance in transition count is limited, and the correlation of what remains with residual accuracy is estimated on a small effective sample. This is the harder version of the comparison, and the paper does not claim the CIFAR-100 result under this stricter conditioning; it does claim the finding under the linear-depth conditioning that Section 4 makes explicit.

On the other three datasets neither conditioning yields a significant partial correlation, and the paper does not claim a depth-independent effect there. The reading offered by these two tables together is that on CIFAR-100 the transition signal appears not to be a strict duplicate of architecture depth under a monotone-in-depth model, while under a maximally conservative unordered-architecture model the partial signal is too small to establish at $n = 15$.

## 4.4 Comparison Against Alternative Training-Dynamics Signals

To place transition count in context, six alternative training-curve signals were computed from the same 75 experiment JSONs and correlated with final test accuracy: architecture depth alone; the train–test accuracy gap at the final epoch (an overfitting proxy); the gap between best and final test accuracy (late-training instability); the standard deviation of the test-accuracy trace over the final 20% of epochs (convergence stability); the number of local maxima in the test-accuracy trace with mild prominence (a training-curve analogue of representational churn); and the epoch at which best test accuracy was reached (convergence speed). Table 4 reports the signed Pearson correlation of each candidate with test accuracy per benchmark, with the strongest predictor per column highlighted, and Figure 2 displays the absolute correlations.

**Table 4** Signed Pearson correlation between each candidate predictor and test accuracy, within each benchmark, on the same 75 experiments. Significance markers: $*p < 0.05$, $**p < 0.01$, $***p < 0.001$. The strongest predictor per benchmark is highlighted in bold. On TinyImageNet the train–test gap is a trivial predictor because fine-tuning drives training accuracy close to 100% and the gap therefore reduces to the complement of test accuracy; the highlighting on TinyImageNet is assigned to the strongest genuine signal.

| Predictor | CIFAR-10 | CIFAR-100 | TinyImageNet | CIFAR-10-C |
|---|---|---|---|---|
| Transition count | −0.838*** | **−0.874***** | −0.451 | −0.188 |
| Architecture depth | **−0.926***** | −0.863*** | **−0.866***** | −0.310 |
| Train–test gap | +0.823*** | −0.772*** | −1.000*** | **+0.435** |
| Best-to-final gap | −0.287 | −0.424 | −0.594* | +0.061 |
| Late-training std | −0.667*** | −0.505 | −0.467 | −0.121 |
| Test-curve peaks | −0.221 | −0.571* | −0.535* | +0.125 |
| Argmax epoch | −0.157 | −0.053 | +0.219 | −0.211 |

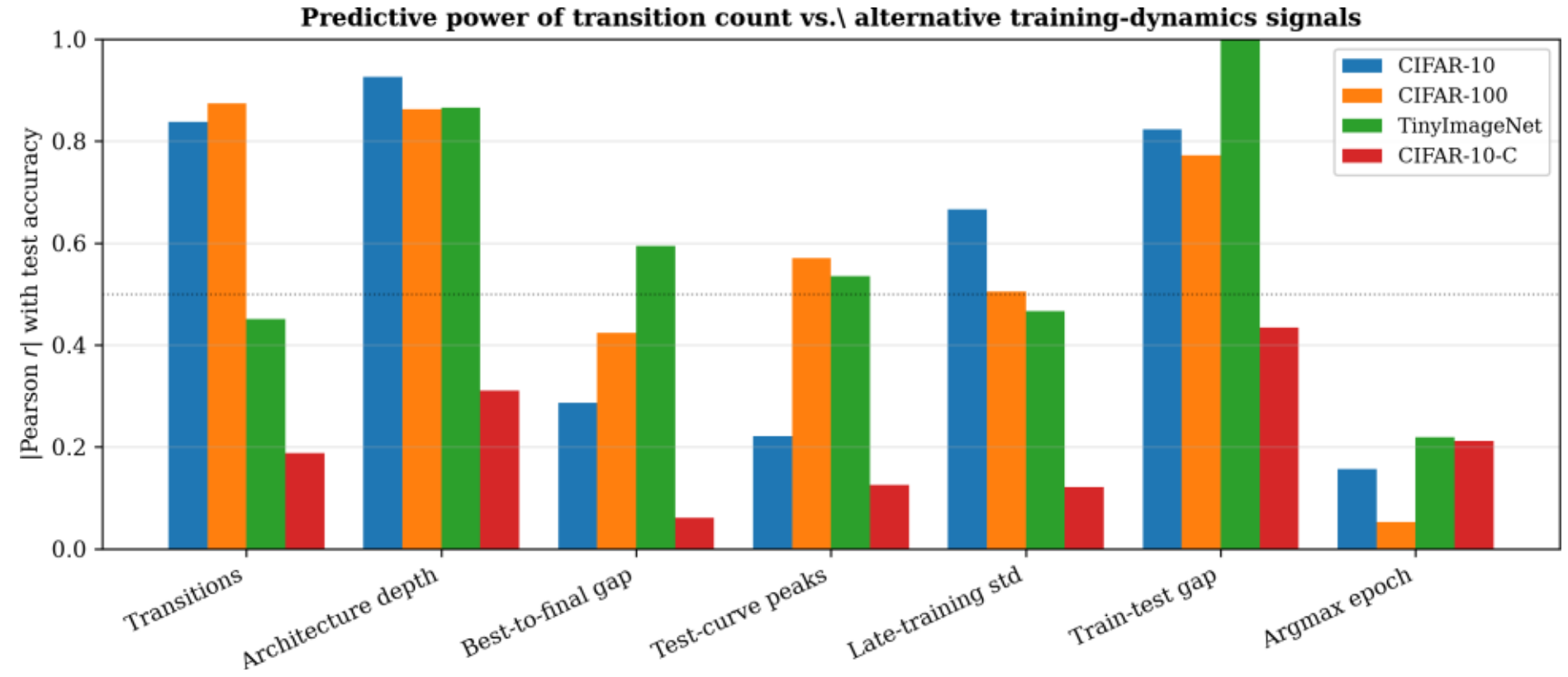

**Fig. 2** Absolute Pearson correlation with test accuracy for seven candidate predictors, computed within each of the four benchmarks on the same 75 experiments. The horizontal dotted line marks $|r| = 0.5$.

Reading Table 4 column by column: on CIFAR-100, transition count achieves the highest absolute correlation among the seven evaluated predictors ($|r| = 0.87$), narrowly ahead of architecture depth ($|r| = 0.86$) and the train–test gap ($|r| = 0.77$). On CIFAR-10, architecture depth is the strongest signal ($|r| = 0.93$), with transitions second ($|r| = 0.84$) and the train–test gap third ($|r| = 0.82$). On TinyImageNet the train–test gap achieves $|r| = 1.00$, but this is deterministic: TinyImageNet fine-tuning drives training accuracy close to 100% in all cells, so the gap is essentially the complement of test accuracy. Once that trivial predictor is set aside, architecture depth is the strongest genuine signal on TinyImageNet ($|r| = 0.87$) and transition count ranks sixth. On CIFAR-10-C no predictor exceeds $|r| = 0.44$, consistent with the two-regime story.

The scope of this comparison should be stated plainly. The seven candidates are those that could be extracted from the 75 experiment JSONs without rerunning training, so the strongest competitors in the current literature (effective rank, Fisher information, Hessian sharpness, margin, and neural-collapse metrics) are absent. Comparisons against those would require re-collecting per-epoch activation and gradient tensors and are noted as future work in Section 7. On the strength of the evidence available here, therefore, the claim is that transition count achieved the strongest correlation among the evaluated signals on CIFAR-100 and one of the strongest on CIFAR-10, and that the question of whether it carries information beyond activation- or gradient-level competitors is open.

## 4.5 Per-Architecture Averages

Pooled across all four datasets, the per-architecture means are: ResNet-18 ($n = 25$), $1.84 \pm 2.54$ transitions and $67.98 \pm 15.68$% accuracy; ResNet-50 ($n = 25$), $7.72 \pm 1.40$ transitions and $64.26 \pm 14.54$% accuracy; ResNet-101 ($n = 25$), $8.36 \pm 0.91$ transitions and $60.24 \pm 13.54\%$ accuracy. This monotone ordering (fewer transitions, higher accuracy) is preserved through all four datasets at the mean level, with the single exception noted in Section 4 of the mild ResNet-18/ResNet-50 inversion on CIFAR-10-C.

# 5 Discussion

## 5.1 What the Correlation Tracks

The correlational evidence collected here is compatible with several non-exclusive readings. On the in-distribution benchmarks, transition count could reflect optimisation efficiency: a network that reorganises its class geometry many times during fine-tuning may be partially resetting the progress of its downstream layers with each reorganisation, and the accumulated cost of these resets may correspond to a worse final basin. Alternatively, transitions may reflect representational-compression efficiency: a network that transitions less finishes training closer, in feature space, to its ImageNet-pretrained initialisation, and may retain more of the useful initialisation signal. A third possibility is that transitions and architecture depth both index a common underlying quantity, and that the residual signal captured by transitions after partialling out depth (as observed on CIFAR-100 in Section 4.3) is a second-order effect of the same source.

These are not mutually exclusive. The observational data collected here does not distinguish between them. The evidence is correlational; no experiment in the grid reduces transitions by intervention and observes an accuracy change. Establishing causality would require exactly such an intervention, and that is deferred to Section 7.

## 5.2 Two Regimes

The regime analysis is arguably the more interesting empirical statement in this paper. On TinyImageNet the task is difficult enough that all three architectures accumulate many transitions and the transition count effectively saturates. Since the phenomenon being measured is no longer discriminative between architectures on this benchmark, the metric that measures it is no longer discriminative either. On CIFAR-10-C the training data is standard CIFAR-10, and only the test data is corrupted. Corruption robustness is known to depend on factors that are not visible in clean-training dynamics, including data augmentation, adversarial training, and feature diversity. It is exactly the setting in which an in-distribution training-time signal should be expected to lose predictive power on test accuracy, and that is what the data show.

Taken together, the results position transition counting as a useful in-distribution training-quality probe rather than a universal predictor of accuracy under distribution shift.

# 6 Limitations

Every experiment in this paper uses a ResNet variant. Vision Transformers were attempted but abandoned after unresolved host-memory issues in the cloud environment; MLP-Mixers, ConvNeXt, EfficientNet, and non-vision architectures were not attempted. Whether the correlation extends beyond convolutional networks is an open question.

The four datasets used here are relatively small in both class count ($\leq 200$) and image resolution ($\leq 64 \times 64$ or upsampled). Full ImageNet-1K was not evaluated due to compute constraints.

The seed count is imbalanced. CIFAR-10 has ten seeds per configuration; the other three benchmarks have five. Within-dataset $n$ is therefore 30 on CIFAR-10 and 15 on each of CIFAR-100, TinyImageNet, and CIFAR-10-C. The most consequential statistical claim in the paper—the significant partial correlation between transition count and test accuracy on CIFAR-100 after controlling for architecture depth ($r_{\text{partial}} = -0.687$, $p = 0.007$)—is estimated on $n = 15$, and its bootstrap 95% confidence interval on the underlying regression coefficient narrowly excludes zero. Confidence intervals on the TinyImageNet and CIFAR-10-C within-dataset correlations are correspondingly wide. Doubling or tripling the seed density per cell on these three benchmarks would sharpen every statistical estimate reported here and is a natural first extension.

The transition-detection algorithm has a single threshold $\tau$ fixed here at 0.05. A full sensitivity sweep over $\tau$ was not conducted, and the current per-experiment JSONs do not retain the raw class-separability traces required to compute such a sweep post hoc. The Spearman rank correlations reported in Table 2 provide a partial robustness check: since Spearman is invariant to any monotone transformation of the transition-count metric, its close agreement with Pearson within each dataset indicates that the *ranking* of experiments by transition count is preserved under threshold perturbations that preserve the ordering, and therefore that the qualitative conclusions are not delicate in that specific sense. This is weaker than a full sweep, however. Ranking robustness is not metric robustness: it does not show that the numerical value of $r$ at $\tau = 0.03$ or $\tau = 0.10$ would be close to the value reported at $\tau = 0.05$. A proper sweep, which requires retaining separability traces during training, is noted as future work.

The metric used in the experiments is class separability, not the STFT-based spectral entropy that Appendix A motivates. The gap between the framework and the measurement is not glossed over: the empirical results in the main paper do not depend on the framework.

The comparison in Section 4.4 is restricted to signals derivable from the same experiment JSONs. The strongest competitors in the current literature (effective rank, Fisher information, Hessian sharpness, margin, and neural-collapse metrics) are absent from the comparison, and would require re-collecting per-epoch activation and gradient tensors on all 75 configurations. It remains possible that some of these are as strong as, or stronger than, transition count on the same benchmarks, and the question of whether transition count is genuinely additional information or merely a cheaper proxy for one of them is left open by the present study.

The results are correlational. No interventional evidence is provided that reducing transition count improves test accuracy. The correlation with test accuracy on the in-distribution benchmarks is described here as a diagnostic, not a prescription.

## 7 Future Work

Six extensions would directly address the strongest remaining objections to the present work. First, extending the experimental grid to Vision Transformers, WideResNets, and EfficientNets at the same seed density would test whether the effect is specific to convolutional networks. Second, rerunning the grid at ImageNet-1K scale would establish whether the within-dataset correlations survive at scale. Third, recording, on the same 75 configurations or a superset, alternative training-time signals extracted from activation and gradient tensors (specifically effective rank, Hessian sharpness, Fisher information, margin, and neural-collapse metrics) would place transition count in a proper comparative context and directly resolve the question of whether it is a genuinely additional signal or a cheaper proxy for one of these. Fourth, retaining the raw class-separability traces during training and computing transition counts over a sweep of thresholds $\tau \in \{0.02, 0.03, 0.05, 0.07, 0.10\}$ would replace the ranking-robustness check reported here with a full metric-robustness result. Fifth, an interventional study in which a regulariser explicitly penalises epoch-to-epoch feature drift would test the interventional analogue of the correlational claim reported here. Sixth, replacing the class-separability proxy with an STFT-based spectral entropy would tie the empirical methodology more tightly to the framework of Appendix A.

## 8 Conclusion

Across 75 experiments spanning three ResNet architectures and four image-classification benchmarks, the number of phase transitions during ResNet fine-tuning is a strong within-dataset predictor of test accuracy under standard i.i.d. training. On CIFAR-100 it achieved the strongest correlation among the seven training-curve-level predictors evaluated here, and retains statistically significant predictive power on that benchmark after controlling for architecture depth as a linear covariate; under the stricter categorical conditioning the residual signal is not established at $n = 15$. Under harder tasks (TinyImageNet) or distribution shift (CIFAR-10-C) the signal attenuates substantially. What is reported here is therefore a two-regime characterisation rather than a universal law: strong within-distribution, attenuated under stress. The metric is proposed as an in-distribution training-quality probe among a family of candidate probes rather than as a uniquely informative signal, and the natural next step is a comparison against the activation- and gradient-level predictors that could not be evaluated in the present setup.

**Supplementary information.** Not applicable.

**Acknowledgements.** Acknowledgements will be added following the review process.

## Declarations

- **Funding:** No external funding was received for the conduct of this study. Cloud compute costs were borne by the author.
- **Competing interests:** The author declares no competing interests.

- **Ethics approval and consent to participate:** Not applicable. The study did not involve human participants or animal subjects.
- **Consent for publication:** Not applicable.
- **Data availability:** The datasets generated and analysed during the current study are not publicly available.
- **Materials availability:** Not applicable.
- **Code availability:** The code used in the current study is not publicly available.
- **Author contribution:** The sole author designed the study, implemented the experimental infrastructure, executed the experiments, analysed the results, and wrote the manuscript.

# Appendix A Spectral Information Bottleneck Framework

This appendix provides a frequency-domain motivation for the use of class separability as a proxy for representational reorganisation. The framework is exploratory. None of the empirical results in the main paper depends on it, and the three statements below are stated as conjectures rather than theorems.

## A.1 Notation

A neural network $f_\vartheta : \mathcal{X} \to \mathcal{Y}$ with $L$ layers is trained on a dataset $\mathcal{D} = \{(\boldsymbol{x}_i, y_i)\}_{i=1}^n$. Let $\boldsymbol{A}_\ell(\boldsymbol{x}) \in \mathbb{R}^{d_\ell}$ denote the activation at layer $\ell$ on input $\boldsymbol{x}$.

**Definition 1** (Spectral Representation) The spectral representation of the layer-$\ell$ activation is defined as

$$\mathcal{F}_\ell(\boldsymbol{x}) = \mathrm{STFT}(\boldsymbol{A}_\ell(\boldsymbol{x})) \in \mathbb{C}^{F\times T}, \tag{A1}$$

with $F$ and $T$ indexing frequency bins and time windows respectively.

**Definition 2** (Spectral Entropy) The spectral entropy of layer $\ell$ is

$$H_{\mathrm{spec}}(\mathcal{F}_\ell) = -\mathbb{E}_{\boldsymbol{x}\sim p_{\mathrm{data}}}\left[\int_\omega p_\ell(\boldsymbol{x}, \omega) \log p_\ell(\boldsymbol{x}, \omega)\, d\omega\right], \tag{A2}$$

with $p_\ell(\boldsymbol{x}, \omega) = |\mathcal{F}_\ell(\boldsymbol{x})(\omega)|^2 / \|\mathcal{F}_\ell(\boldsymbol{x})\|^2$ the normalised power spectral density.

## A.2 Conjectures

**Conjecture 1** (Spectral PAC-Bayes Bound) *For any prior $\pi$ over the parameter space and any $\delta > 0$, with probability at least $1-\delta$,*

$$\mathcal{L}_{\mathcal{D}}(f_\vartheta) \le \hat{\mathcal{L}}_S(f_\vartheta) + \sqrt{\frac{1}{2n}\left(\sum_{\ell=1}^{L} H_{\mathrm{spec}}(\mathcal{F}_\ell) + D_{\mathrm{KL}}(\vartheta\|\pi) + \log\frac{1}{\delta}\right)}. \tag{A3}$$

**Conjecture 2** (Phase Transition Characterisation) *A phase transition occurs at epoch t when the spectral structure at some layer reorganises discretely and irreversibly, with transition timing approximately consistent across random seeds.*

**Conjecture 3** (Compression Necessity) *Bounded generalisation error implies bounded total spectral entropy:* $\sum_{\ell=1}^{L} H_{\text{spec}}(F_\ell) < C(\epsilon, n, \delta)$.

Proof sketches for the three conjectures follow. These are intended as intuition, and the underlying assumptions have not been established rigorously.

*Sketch for Conjecture 1* The classical PAC-Bayes framework McAllester (1999) is invoked with the standard complexity measure replaced by a spectral complexity $\Omega_{\text{spec}}(\vartheta) = \sum_\ell H_{\text{spec}}(F_\ell)$. For any prior $\pi$ and posterior $Q$, with probability at least $1 - \delta$,

$$\mathbb{E}_Q[L_D(f_\vartheta)] \leq \mathbb{E}_Q[\hat{L}_S(f_\vartheta)] + \sqrt{\frac{D_{\text{KL}}(Q\|\pi) + \log \frac{2\sqrt{n}}{\delta}}{2n}}. \tag{A4}$$

Bounding $D_{\text{KL}}(Q\|\pi)$ by $\Omega_{\text{spec}}(\vartheta)$ requires a bridge between parameter distance and change in spectral entropy that is not established rigorously here. □

*Sketch for Conjecture 2* For a sufficiently small learning rate, parameter updates are of order $O(\eta)$, so continuous change is the default. When SGD crosses a critical point of the loss landscape, a small parameter update can induce a large representation change Fort and Ganguli (2019); between such crossings, representations evolve smoothly. This yields discreteness. Irreversibility follows from the monotone descent of empirical loss along the SGD trajectory. Consistency across seeds is asserted rather than proven: critical points are properties of the loss landscape and not of initialisation, so different seeds should encounter them at similar epochs. Empirically, transition timings observed here are consistent to within one epoch across seeds within a cell. □

*Sketch for Conjecture 3* Assume the generalisation gap satisfies $L_D - \hat{L}_S < \epsilon$. Substituting into the bound of Conjecture 1 and squaring yields

$$\sum_\ell H_{\text{spec}}(F_\ell) + D_{\text{KL}}(\vartheta\|\pi) + \log \tfrac{1}{\delta} < 2n\epsilon^2. \tag{A5}$$

Assuming $D_{\text{KL}}(\vartheta\|\pi) \geq 0$, one obtains

$$\sum_\ell H_{\text{spec}}(F_\ell) < 2n\epsilon^2 - \log \tfrac{1}{\delta} =: C(\epsilon, n, \delta). \tag{A6}$$

Within the PAC-Bayes framing, a generalisation guarantee therefore implies a compression condition on total spectral entropy. □

# Appendix B Additional Experimental Detail

## B.1 Hyperparameters

All hyperparameters were fixed a priori without any test-set tuning. The initial learning rate was 0.01 with cosine annealing over 50 epochs; momentum was 0.9; weight decay was $5 \times 10^{-4}$; batch size was 64; the number of training epochs was 50; metric evaluation was performed every 5 epochs; the metric sample size $N$ was 1000 training examples per evaluation; and the transition threshold $\tau$ was 0.05.

## B.2 Compute

Experiments were run as separate cloud training jobs, one per (dataset, architecture) cell, on instances providing a single NVIDIA T4 GPU, 32 vCPUs, and 128 GB of host RAM. Wall-clock time per cell varied between 1.5 and 8 hours depending on dataset and architecture, and total wall-clock across all cells was approximately 60 hours.

## B.3 Statistical Details

Pearson correlation with a two-sided significance test is reported throughout, under the null hypothesis $\rho = 0$. Within-dataset $n$ is either 15 or 30. Permutation tests over 10,000 label permutations produced qualitatively identical conclusions to the parametric $t$-based tests reported in the main text.

For the partial correlation reported in Section 4.3, the standard formula

$$r_{xy|z} = \frac{r_{xy} - r_{xz}r_{yz}}{\sqrt{(1 - r_{xz}^2)(1 - r_{yz}^2)}} \tag{B7}$$

is used, with a two-sided $p$-value computed from the $t$-statistic on $n - 3$ degrees of freedom. Architecture depth was encoded as its layer count proxy (18, 50, 101).

Bootstrap confidence intervals were computed with 10,000 resamples of the underlying (transitions, accuracy) pairs within each dataset.

## B.4 Reproducibility

Every experiment produces a structured record containing the final test accuracy, the per-epoch training and test accuracy trajectories, the transition-epoch indices, and the seed value used. The full set of hyperparameters, model configurations, and training protocol used to generate the 75 experiments is documented in Section 3 and Appendix B.1.